\documentclass[conference,letterpaper]{IEEEtran}
\IEEEoverridecommandlockouts
\usepackage{graphicx}
\usepackage{bm}
\usepackage{multirow}
\usepackage{amssymb}
\usepackage{amstext}
\usepackage{amsmath}
\usepackage{amsthm}
\usepackage{multicol}
\usepackage{hyperref}

\usepackage{todonotes}
\usepackage{regexpatch}
\makeatletter
\xpatchcmd{\@todo}{\setkeys{todonotes}{#1}}{\setkeys{todonotes}{inline,#1}}{}{}
\makeatother

\usepackage{subfigure}

\newcommand{\matrx}[1]{\bm{#1}}
\newcommand{\vectr}[1]{\bm{#1}}
\newcommand{\randvar}[1]{\textnormal{#1}}
\newcommand{\E}{\mathbb{E}}
\newcommand{\KL}{D_{\mathrm{KL}}}

\newcommand{\recon}{$\ell_{AE}$}
\newcommand{\cross}{$\ell_{H}$}
\newcommand{\kmeans}{$k$-means}
\newcommand{\vx}{\ensuremath{\boldsymbol{x}}}

\DeclareMathOperator*{\argmax}{\arg\!\max}

\begin{document}
\title{Evidence Transfer for Improving Clustering Tasks Using External Categorical Evidence}
\author{
    \IEEEauthorblockN{Athanasios Davvetas, Iraklis A. Klampanos, Vangelis Karkaletsis}
\IEEEauthorblockA{\textit{Institute of Informatics and Telecommunications} \\
    \textit{National Centre for Scientific Research ``Demokritos''}\\
    Agia Paraskevi, Athens, Greece \\
\{tdavvetas, iaklampanos, vangelis\}@iit.demokritos.gr}
}

\maketitle

\begin{abstract}
    In this paper we introduce evidence transfer for clustering, a deep learning method that can incrementally manipulate the latent representations of an autoencoder, according to external categorical evidence, in order to improve a clustering outcome. 
    By evidence transfer we define the process by which the categorical outcome of an external, auxiliary task is exploited to improve a primary task, in this case representation learning for clustering. 
    Our proposed method makes no assumptions regarding the categorical evidence presented, nor the structure of the latent space. We compare our method, against the baseline solution by performing k-means clustering before and after its deployment. Experiments with three different kinds of evidence show that our method effectively manipulates the latent representations when introduced with real corresponding evidence, while remaining robust when presented with low quality evidence. 
\end{abstract}


\section{Introduction}
\label{sec:intro}

Over the last few years an increasing number of studies have proposed using multi modal datasets, or external evidence, as a way to increase performance. Among other tasks, question answering \cite{Savenkov2017}, action recognition \cite{Park2016} and social media inference \cite{Li2015} have made use of external information to learn increasingly meaningful representations. These representations are a product of co-learning the primary and the external data through a common objective. In the context of supervised learning, this learning process depends on the availability of external data as well as on their relation to the primary dataset.

However, in practice the availability of external data is either not guaranteed, or we may observe the outcome of external processes without having explicit access to the corresponding dataset. 
For instance, suppose we want to cluster towns according to their observed weather (primary task). At the same time we observe a grouping of geographical regions (which may include more than one town) according to rainfall data (auxiliary task). The outcome of the auxiliary task clearly relates to the primary task, however (a) the actual relationship is unknown, i.e. we do not know the extent to which rainfall can predict the overall weather of an area; (b) we do not have access to the data which led to the auxiliary tasks (in this case, the rainfall-based outcome); 
and (c) even though data items (and consequently task outcomes) in the two tasks will be related, the cardinality of this relationship is unknown (e.g. in this example each region may contain multiple towns). Here, we consider the auxiliary categorical outcome as external evidence. Evidence transfer is then the process of influencing the latent representations of a dataset to improve a primary task.

In this paper we propose a general framework that uses external categorical evidence when available to improve unsupervised learning, and in particular clustering. Using autoencoders that learn the primary dataset distribution, we learn a latent space that can be later manipulated to reflect new external evidence.

We present a general evidence transfer method for combining multiple sources of external evidence to improve the outcome of clustering tasks. Our method makes no assumptions regarding the quality, source or availability of external information. It aids clustering tasks by learning augmented latent representations that are separated according to external categorical evidence. By manipulating the latent space, we increase the effectiveness of clustering algorithms that rely on linear distance metrics, such as \kmeans{}. Our method is effective, robust when presented with low quality of additional evidence, and modular as it can be incrementally applied to new pieces of evidence.

\section{Methodology}
\label{sec:method}
In this section we define the problem of combining external evidence in a primary clustering task, we introduce appropriate fitness criteria and propose an evidence transfer method that satisfies these  criteria.

\subsection{Problem Statement}

Consider the task of clustering a dataset $\bm{X} = \{\vx^{(1)}, \vx^{(2)}, \dots, \vx^{(N)} \}$.  
Clustering $\bm{X}$ yields a set of memberships $\bm{C} = \{\bm{c}^{(1)}, \bm{c}^{(2)}, \dots, \bm{c}^{(N)}\}$, where $\bm{c}^{(i)}$ can be modelled as the categorical probability distribution of $\bm{x}^{(i)}$ over the target classes, $p(c|\bm{x})$. For the case of hard cluster assignment, one would eventually assign $\bm{x}^{(i)}$ to cluster $\argmax \bm{c}^{(i)}$. In this paper $\matrx{X}$ denotes the primary dataset and clustering $\bm{X}$ is the primary task. 

External evidence $\bm{V}=\{\bm{v}^{(1)}, \bm{v}^{(2)}, \dots, \bm{v}^{(M)}\}$ is the set of outcomes of an auxiliary task carried out either on the primary dataset $\matrx{X}$ or on some unknown secondary dataset. Similar to the primary task, each $\bm{v}^{(j)}$ can be seen as the categorical distribution of a subset of $\bm{X}$ over the target classes of the auxiliary task, with the most straightforward case being that $M=N$ and that there is a one-to-one correspondence between the elements of $\bm{X}$ and $\bm{V}$. There may exist multiple sources of external evidence yielding observable membership outcomes $\bm{\mathsf{V}} = \{\bm{V}_1, \bm{V}_2, \dots, \bm{V}_K \}$.

Our objective is to use external evidence $\bm{\mathsf{V}}$ to improve accuracy by reducing uncertainty in the primary clustering task. 
In any clustering task there are data samples that the clustering algorithm is not able to distinguish with high certainty. Clustering on latent representations of $\bm{X}$ generally leads to better results due to their increased linear separability in the latent space learned \cite{Bengio}. By allowing external evidence $\bm{\mathsf{V}}$ to also influence the learning process, we posit that we can further improve linear separability, therefore achieving increased certainty in the primary task.


In this work we do not take into account the auxiliary datasets directly, only external categorical evidence produced on an unseen dataset by an unseen process. It follows that the proposed method makes no assumptions regarding the relation of the external evidence to the primary dataset. The only assumption made is that each external evidence is somehow related to the primary dataset but any mapping $f(\bm{X}) \mapsto \bm{V}_{K}$ is unknown or too complex. 

Methods that attempt to improve 
a task via evidence transfer
should be at least effective when presented with helpful external evidence, and robust against low quality or irrelevant evidence. In practice, sources of evidence may not be known and available at the beginning of evidence transfer, and therefore methods should also be modular in order to allow incrementally improving representations. More specifically, we define the following fitness criteria for evidence transfer.
\begin{enumerate}
    \item \textbf{Effectiveness}: In the case that the evidence corresponds to a meaningful relation between itself and the primary dataset, this should be discovered and utilised to reduce uncertainty in the latent space. Meaningful relations are characterised by consistency on the outcome of the auxiliary tasks. Intuitively, introducing more than one sources of consistent evidence should lead to more effective performance than using a single source of evidence.
    \item \textbf{Robustness}: Since the mapping $f(\bm{X}) \mapsto \bm{V}_{K}$ is unknown, there may be evidence that does not contribute meaningful information to the primary task, i.e. that it does not contribute to a meaningful separation of the latent representations of $\bm{X}$. For example, in cases where $\bm{V} \sim \mathcal{U}$ distribution or in cases where the evidence is consistent but it is introduced in a non corresponding order. The algorithm should be able to distinguish this evidence as low quality evidence and be able to reject it without making significant changes in the latent space, therefore maintaining its prior effectiveness.
    \item \textbf{Modularity}: The method should not require complete re-training in view of additional evidence. For instance, the proposed method includes evidence as a fine tune step that augments the baseline representations. The added step should not disrupt the latent space in such way that will lead to changes in the original objective. The transformations that take place during the finetune step, should be restricted by the original objective of the baseline solution.
\end{enumerate}

In summary, \textit{modular} and \textit{robust} solutions behave well against low quality evidence, while \textit{effective} solutions reduce uncertainty in the latent space, leading to better performance.

\subsection{Dealing with external evidence}

In order to satisfy the above criteria we consider the minimization of cross entropy as an appropriate objective. Cross entropy is an asymmetrical metric that involves the entropy of the ``true'' distribution and its divergence to an auxiliary distribution (Equation \ref{eq:basiccrossent}). Considering the external evidence as the ``true'' distribution and the latent space as the ``auxiliary'' distribution, then cross entropy quantifies the uncertainty of evidence distribution, as well as, its relation to the latent space. As a task outcome, evidence distribution is considered as fixed and therefore its entropy is constant. On the other hand, the distribution of the latent space belongs to parametric families that involve the trainable parameters of the neural network.
\begin{equation}\label{eq:basiccrossent}
    H(\randvar{P}, \randvar{Q}) = \E_\randvar{P}[-\log \randvar{Q}] = H(\randvar{P}) + \KL(\randvar{P} \Vert \randvar{Q})
\end{equation}

We use the cross entropy to shift these parameters into reducing the divergence between the evidence distribution and the latent space. In cases where evidence correlates with the latent space, their divergence is minimized and therefore satisfying the effectiveness criterion. In cases where evidence can not be correlated with the latent space, their divergence converges to high values that affect the latent space less and less with each epoch.

Since the relation between the evidence and the primary dataset is unknown, we do not introduce the evidence samples in their raw format. We use a biased additional autoencoder with a single hidden layer to upscale or downscale the width of the latent evidence samples, to correspond with the width of the inner hidden layer of the primary autoencoder. We use the term biased, due to not allowing the evidence autoencoder to generalize over the input dataset. We train the evidence autoencoder with low number of epochs to act as an identity function. 

We use latent categorical representations of the biased evidence autoencoders, denoted as $\bm{Z}_{\mathsf{{V}}}$, since the identity bias forces the evidence autoencoder to produce the same latent space distribution for both white noise evidence and inconsistent evidence. In both cases, the latent evidence samples approximate a uniform distribution which results in high cross entropy that converges to constant loss.

\subsection{Evidence transfer}

Our method is a sequence of two steps, the \textit{initialization} and \textit{evidence transfer} steps.
During the initialization step we introduce a baseline clustering method that we later finetune using external additional evidence. In order to initialize the latent space of the baseline solution, we train an autoencoder using the standard MSE loss,  \recon{}: 
\begin{equation}\label{eq:aerecon}
    \ell_{AE} = \mathcal{L}(\matrx{\tilde{X}}, \matrx{X'}) = \frac{1}{N} \sum_{i=1}^{N} (\vectr{\tilde{x}}^{(i)}-\vectr{x'}^{(i)})^2
\end{equation}
Generative types of autoencoder such as denoising autoencoders, variational autoencoders \cite{Kingma2013} or adversarial autoencoders \cite{Makhzani2015} are fit for the initialization of the latent space. Generative autoencoders approximate a latent space distribution that is close to the true underlying data generation distribution. In our experiments, we use denoising autoencoders that maximize the expected log-likelihood of dataset $\bm{X}$ given corrupted dataset $\bm{\tilde{X}}$ (by minimizing Equation \ref{eq:likelihood}, as defined in \cite{Bengio2013}), with expectation taken over the joint data-generating distribution. 

\begin{equation}\label{eq:likelihood}
    \mathcal{L}(\theta) = -\E_{P(\bm{X},\bm{\tilde{X}})}{[ \log P_\theta(\bm{X}|\bm{\tilde{X}})]}
\end{equation}

When the training of the autoencoder is done, we use the \kmeans{} algorithm on the initial latent representations. We introduce this method as a baseline solution to the clustering problem. Before we proceed to the evidence transfer step, we also train an additional evidence autoencoder for each source of evidence, to produce latent categorical samples $\bm{Z}_{\mathsf{{V}}}$ as part of the initialization step, using Equation \ref{eq:eviaerecon}.

\begin{equation}\label{eq:eviaerecon}
    \ell_{EviAE} = \mathcal{L}(\bm{V}_K, \bm{V}'_K) = \frac{1}{M} \sum_{i=1}^{M} (\bm{v}_K^{(i)}-\bm{v}_K'^{(i)})^2
\end{equation}

\begin{equation}\label{eq:crossent}
    \ell_{H} = \frac{1}{K} \sum_{j=1}^{K}H(\randvar{Z}_{V_j}, \randvar{Q}_j)
\end{equation}

\begin{equation}\label{eq:evitram}
    \ell_{EviTRAM} = \ell_{AE} + \lambda * \ell_{H}
\end{equation}

Objective \cross{} minimizes the mean cross entropy between each additional evidence sources and predictors $Q$. Objective \recon{} restricts the latent space to preserve its baseline structure, meaning that latent samples are able to perform their original task, which is the reconstruction of primary dataset samples. We jointly minimize both \recon{} and \cross{} by minimizing Equation \ref{eq:evitram} that involves both losses, using hyperparameter $\lambda$ as the coefficient of \cross{} loss.
By jointly optimizing both tasks we approach the maximization of the expected log-likelihood by using evidence informed parameters $\theta$.

\begin{figure*}
  \centering
  \subfigure[Primary Autoencoder with additional layers (Stacked)]{\includegraphics[width=.6\textwidth]{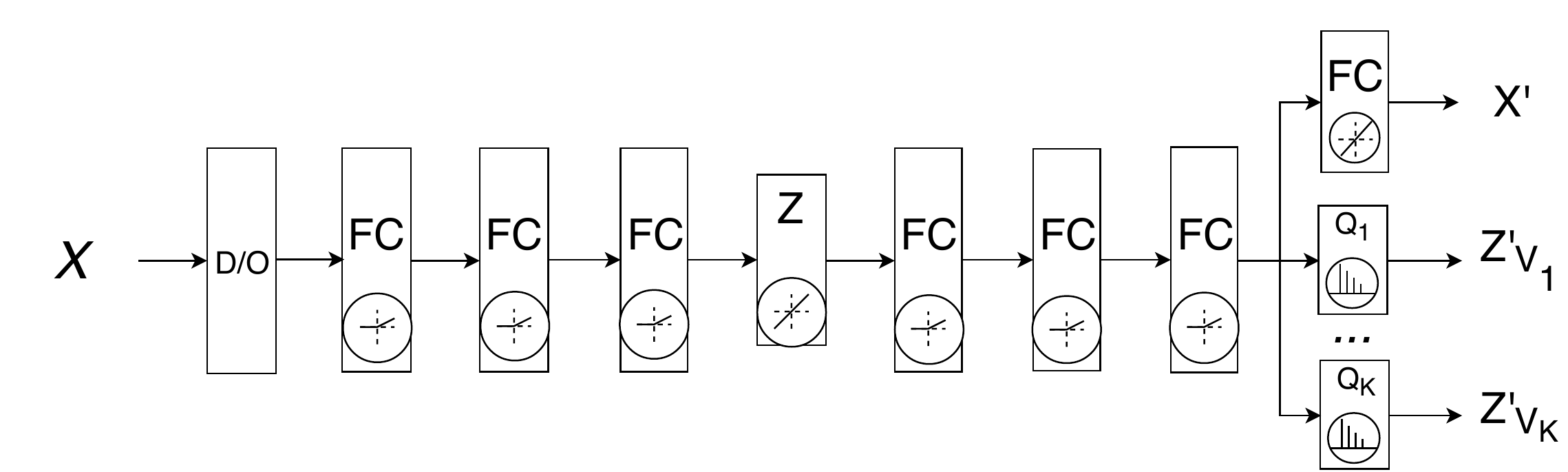}} 
  \subfigure[Primary Autoencoder with additional layers (Convolutional)]{\includegraphics[width=.9\textwidth]{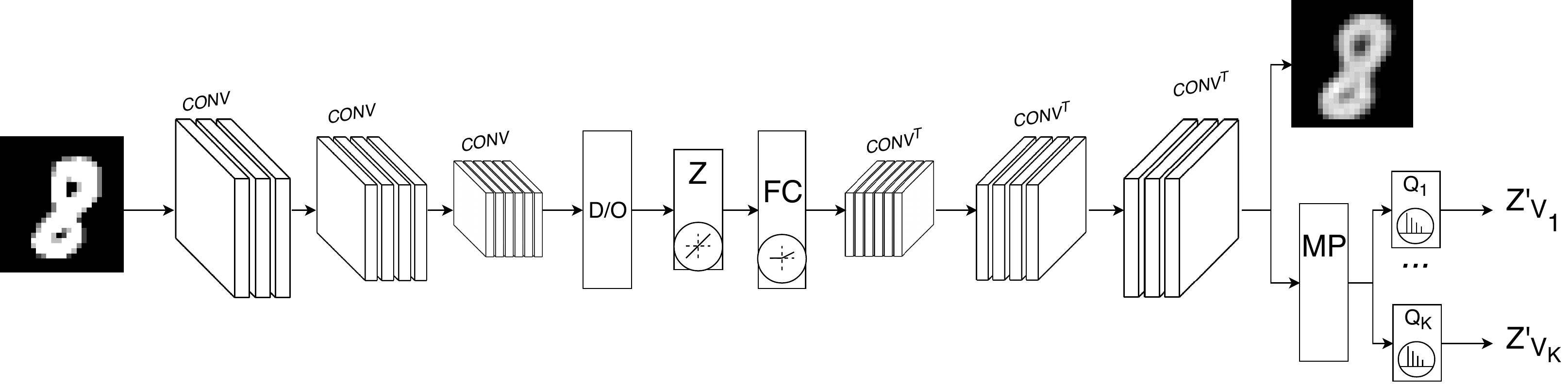}} 
  \subfigure[Evidence Autoencoder (Auxiliary)]{\includegraphics[width=0.3\textwidth]{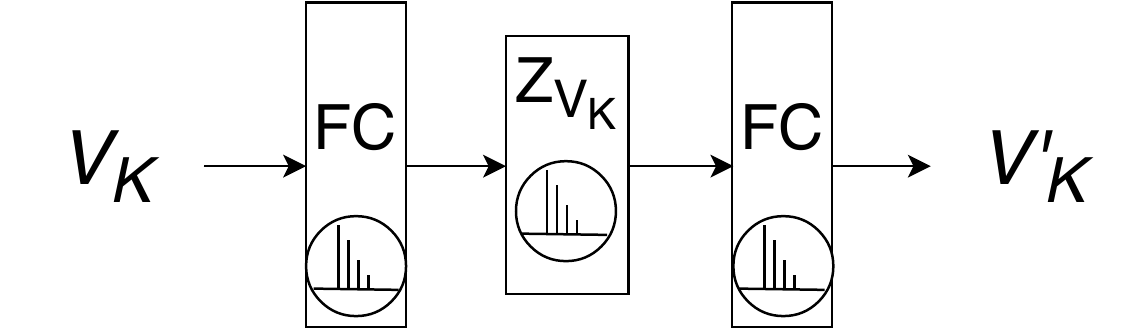}}
  \caption{Neural network configurations used by evidence transfer method. Figure (a) and (b) depict the neural network configuration of the primary task, which is the initial baseline solution along with the additional layers need to manipulate the latent space. We showcase the Stacked Denoising Autoencoder version (Figure (a)) that was used for the experiments in CIFAR, 20newsgroups and REUTERS-100k. For the MNIST dataset, we use a Convolutional Autoencoder (Figure (b)), the autoencoder topology is similar to the proposed Convolutional Autoencoder in DCEC. The Stacked Denoising Autoencoder has widths of d-500-500-200-10, as seen in DEC. Figure (c) depicts the topology of the evidence autoencoder where we acquire $Z_{V_{k}}$ latent representations that are used to during the evidence transfer method. The widths of the fully connected layers in the evidence autoencoder depend on the widths of each $V_k$ evidence and $Z$ latent space width.}
  \label{fig:conf}
\end{figure*}

We use additional layers (one for each source of evidence) in the output of the autoencoder in order to predict the latent categorical variables $\bm{Z}_{\mathsf{{V}}}$. As depicted in Figure \ref{fig:conf}. Opposing to directly manipulating the latent space, predictors $Q$ adjust their weights depending on the quality of the evidence. In cases of low quality evidence, their weights decay and the joint minimization of Equation \ref{eq:evitram}, is achieved by minimizing \recon{}.


\section{Evaluation and Results}
\label{sec:eval}

For the purpose of evaluating our solution we tried three different qualities of evidence (real corresponding evidence, random values / white noise, real corresponding evidence introduced in random order) and three different quantities of evidence (single, double, triple). The criteria of fitness of our solution are both the effectiveness and robustness. Random index evidence, is essentially real corresponding evidence. We introduce it in an non corresponding order to evaluate the robustness of our solution on inconsistent evidence.

\subsection{Datasets and Metrics}
We briefly introduce the datasets and the preprocess techniques that were use in our experiments. 
\begin{itemize}
    \item \textbf{MNIST}: The MNIST dataset consists of 70000 images of handwritten digits. Each 28 x 28 image is reshaped into a single vector with 784 features.
    \item \textbf{CIFAR-10}: CIFAR-10 contains 60000 32x32 colour images of 10 classes. In a similar manner as the experiments in VADE \cite{Jiang2017} and DEC \cite{Xie2016}, we do not cluster the raw images. We use feature vectors acquired by a pretrained VGG-16 network on ImageNet. We use the output of the first dense layer of VGG-16 as input to our configuration, each image is transformed to a single vector of 4096 features. 
    \item \textbf{20 Newsgroups}: A dataset of 20000 newsgroup documents, for our experiments we use features acquired from a pretrained word2vec model \cite{Mikolov} on Google news corpus. We acquire a 300 dimensional vector for each word. After the preprocess we acquire 18282 documents. To represent each document we use the mean of its word embeddings. 
    \item \textbf{Reuters}: Reuters Corpus Volume I \cite{Lewis2004} contains 804414 documents of 103 categories. For our experiments we use a subset of 96933 documents of 10 sub categories. In the same manner as DEC, we compute tf-idf features on the 2000 most frequent word stems.
\end{itemize}

To evaluate the effectiveness and robustness of each experiment we use the unsupervised clustering accuracy (ACC) and the normalized mutual information score (NMI) metrics. The unsupervised clustering accuracy was introduced in DEC. Both metrics are used in the evaluation of latent representation clustering frameworks such as DEC, VADE, DCEC \cite{Guo2017}, DEPICT \cite{Dizaji2017}, JULE \cite{Yang2016}, etc.

\subsection{Effectiveness and Robustness}

We evaluate our solution based on the results of our experiments when introducing one, two and three sources of evidence (Tables \ref{tab:imagesets}, \ref{tab:textsets} and \ref{tab:triple}). In all cases where a corresponding source of evidence is present, our solution is able to effectively utilize it, leading to increase of the unsupervised clustering accuracy and normalized mutual information score. The effectiveness criterion is successfully satisfied, considering that the gain in effectiveness is scalable with the amount of corresponding evidence sources. The robustness criterion is also satisfied since there is no significant loss in cases where we introduce any source of low quality evidence.

\begin{table*}[h]
    \caption{For our experiments in both MNIST and CIFAR, we report the average of 4 runs for each evidence configuration. $W$ indicates the width of each evidence vector. For MNIST, real evidence with width 3 represents the $y \mod 3$ relation ($y$ being the digit label), while evidence with width equal to 4 corresponds to the relation $hash(y) \mod 4$. Real evidence of 10 width, is the full labelset of MNIST. For CIFAR-10, real evidence does not correspond to any notable relations. Width 3 real evidence separates the samples into three categories: vehicles, pets and wild animals. Width 4 real evidence expands the pets category into additional two group of two classes. 10 width evidence corresponds to the labelset of CIFAR-10. }
\label{tab:imagesets} \centering %

\subtable[MNIST]{
    \begin{tabular}{|c|c|c|}
    \hline 
    \multirow{1}{*}{} & ACC (\%) & NMI (\%)\tabularnewline
    \hline 
    Baseline & 82.03 & 76.25\tabularnewline
    \hline 
    Real evidence (w: 3) & 95.57 (+13.54) & 89.59 (+13.34)\tabularnewline
    \hline 
    Real evidence (w: 10) & 96.71 (+14.68) & 91.77 (+15.52)\tabularnewline
    \hline 
    White noise (w: 3) & 82.32 (+0.29) & 76.40 (+0.14)\tabularnewline
    \hline 
    White noise (w: 10) & 82.32 (+0.29) & 76.40 (+0.14)\tabularnewline
    \hline 
    Random index (w: 3) & 82.16 (+0.13) & 76.29 (+0.04)\tabularnewline
    \hline 
    Random index (w: 10) & 82.34 (+0.32) & 76.43 (+0.18)\tabularnewline
    \hline 
    Real (w: 3) + Real (w: 4) & 97.72 (+15.69) & 93.93 (+17.68)\tabularnewline
    \hline 
    Noise (w: 3) + Noise (w: 10) & 82.20 (+0.17) & 76.38 (+0.13)\tabularnewline
    \hline 
    Real (w: 3) + Noise (w: 3) & 95.52 (+13.50) & 89.50 (+13.25)\tabularnewline
    \hline 
    \end{tabular}
}
\subtable[CIFAR-10]{
    \begin{tabular}{|c|c|c|}
    \hline 
    \multirow{1}{*}{} & ACC (\%) & NMI (\%)\tabularnewline
    \hline 
    Baseline & 22.79 & 13.44\tabularnewline
    \hline 
    Real evidence (w: 3) & 37.34 (+14.56) & 46.24 (+32.80)\tabularnewline
    \hline 
    Real evidence (w: 10) & 91.97 (+69.18) & 83.06 (+69.62)\tabularnewline
    \hline 
    White noise (w: 3) & 24.62 (+1.83) & 14.66 (+1.22)\tabularnewline
    \hline 
    White noise (w: 10) & 24.61 (+1.82) & 14.56 (+1.12)\tabularnewline
    \hline 
    Random index (w: 3) & 26.18 (+3.39) & 15.35 (+1.91)\tabularnewline
    \hline 
    Random index (w: 10) & 26.01 (+3.22) & 15.08 (+1.63)\tabularnewline
    \hline 
    Real (w: 3) + Real (w: 4) & 52.86 (+30.07) & 61.44 (+48.00)\tabularnewline
    \hline 
    Noise (w: 3) + Noise (w: 10) & 25.00 (+2.22) & 14.80 (+1.35)\tabularnewline
    \hline 
    Real (w: 3) + Noise (w: 3) & 36.97 (+14.18) & 46.22 (+32.78)\tabularnewline
    \hline 
    \end{tabular}
}
\end{table*}

\begin{table*}[h]
    \caption{For our experiments in both 20 Newsgroups and REUTERS-100k, we report the average of 4 runs for each evidence configuration. $W$ indicates the width of each evidence vector. For the 20 Newsgroups dataset, we used its natural structure as evidence, the 20 original labels are accompanied by a prefix that indicates a root category. We divided the labelset into ``comp(uters).'', ``rec(reational)'', ``sci(ence)'', ``talk'' and ``misc.'', to produce a 5 width real evidence. To produce 6 width real evidence, we divided the labelset into 6 classes namely ``sport'', ``politics'', ``religion'', ``vehicles'', ``systems'' and ``science''. 20 width evidence corresponds to the labelset of the full dataset. For REUTERS-100k, width 4 real evidence represents the 4 root categories, while 5 width real evidence is a simple re-categorization of the 10 labels into 5 groups of two classes. Width 10 corresponds to the labelset of 10 sub categories (C15, C151, GPOL, GSPO, GDIP, E51, M11, M14, E21, E41). }
\label{tab:textsets} \centering %

\subfigure[20 Newsgroups]{
    \begin{tabular}{|c|c|c|}
    \hline 
    \multirow{1}{*}{} & ACC (\%) & NMI (\%)\tabularnewline
    \hline 
    Baseline & 21.19 & 25.01\tabularnewline
    \hline 
    Real evidence (w: 5) & 34.18 (+12.99) & 57.35 (+32.34)\tabularnewline
    \hline 
    Real evidence (w: 20) & 88.90 (+67.71) & 90.01 (+65.00)\tabularnewline
    \hline 
    White noise (w: 3) & 22.36 (+1.17) & 25.49 (+0.49)\tabularnewline
    \hline 
    White noise (w: 10) & 22.46 (+1.27) & 26.11 (+1.10)\tabularnewline
    \hline 
    Random index (w: 5) & 21.77 (+0.58) & 25.32 (+0.32)\tabularnewline
    \hline 
    Random index (w: 20) & 22.40 (+1.21) & 25.54 (+0.53)\tabularnewline
    \hline 
    Real (w: 5) + Real (w: 6) & 46.19 (+25.00) & 68.31 (+43.30)\tabularnewline
    \hline 
    Noise (w: 3) + Noise (w: 10) & 22.89 (+1.70) & 26.35 (+1.34)\tabularnewline
    \hline 
    Real (w: 5) + Noise (w: 3) & 31.41 (+10.22) & 54.24 (+29.24)\tabularnewline
    \hline 
    \end{tabular}
}
\subfigure[REUTERS-100k]{
    \begin{tabular}{|c|c|c|}
    \hline 
    \multirow{1}{*}{} & ACC (\%) & NMI (\%)\tabularnewline
    \hline 
    Baseline & 41.12 & 32.72\tabularnewline
    \hline 
    Real evidence (w: 4) & 43.34 (+2.22) & 36.24 (+3.52)\tabularnewline
    \hline 
    Real evidence (w: 10) & 48.27 (+7.15) & 41.23 (+8.51)\tabularnewline
    \hline 
    White noise (w: 3) & 41.42 (+0.30) & 32.77 (+0.05)\tabularnewline
    \hline 
    White noise (w: 10) & 41.38 (+0.26) & 32.74 (+0.03)\tabularnewline
    \hline 
    Random index (w: 4) & 41.37 (+0.25) & 32.82 (+0.10)\tabularnewline
    \hline 
    Random index (w: 10) & 41.38 (+0.26) & 32.68 (-0.03)\tabularnewline
    \hline 
    Real (w: 4) + Real (w: 5) & 50.54 (+9.42) & 41.81 (+9.10)\tabularnewline
    \hline 
    Noise (w: 3) + Noise (w: 10) & 41.16 (+0.04) & 32.65 (-0.06)\tabularnewline
    \hline 
    Real(w: 4) + Noise (w: 3) & 43.44 (+2.32) & 36.29 (+3.57)\tabularnewline
    \hline 
    \end{tabular}
}
\end{table*}

\begin{table*}[h]
    \caption{In this table we report our experiment of introducing three sources of evidence in the CIFAR-10 dataset. We report the average of 4 runs for each evidence configuration. $W$ indicates the width of each evidence vector. We used the evidence introduced in table \ref{tab:imagesets}, plus one additional random value evidence with width 5 and one additional width 5 real evidence that turns the labelset into 5 groups of 2 two classes.}
\label{tab:triple} \centering %

\begin{tabular}{|c|c|c|}
\hline 
\multirow{1}{*}{} & ACC (\%) & NMI (\%)\tabularnewline
\hline 
Baseline & 22.79 & 13.44\tabularnewline
\hline 
Real (w: 3) + Real (w: 4) + Real (w: 5) & 64.75 (+41.96) & 74.23 (+60.79)\tabularnewline
\hline 
Real (w: 3) + Real (w: 4) + Noise (w: 3) & 53.04 (+30.26) & 61.74 (+48.30)\tabularnewline
\hline 
Real (w: 3) + Noise (w: 3) + Noise (w: 10) & 36.67 (+13.89) & 46.21 (+32.77)\tabularnewline
\hline 
Real (w: 3) + Real (w: 5) + Noise (w: 3) & 60.56 (+37.77) & 71.39 (+57.94)\tabularnewline
\hline 
Real (w: 3) + Noise (w: 3) + Noise (w: 10) & 44.68 (+21.89) & 54.37 (+40.92)\tabularnewline
\hline 
Real (w: 4) + Real (w: 5) + Noise (w: 3) & 63.42 (+40.63) & 77.16 (+63.72)\tabularnewline
\hline 
Real (w: 5) + Noise (w: 3) + Noise (w: 10) & 62.49 (+39.70) & 65.58 (+52.14)\tabularnewline
\hline 
Noise (w: 3) + Noise (w: 10) + Noise (w: 5) & 25.21 (+2.43) & 14.90 (+1.46)\tabularnewline
\hline 
\end{tabular}
\end{table*}

\begin{figure}
  \centering
  \includegraphics[width=.35\textwidth]{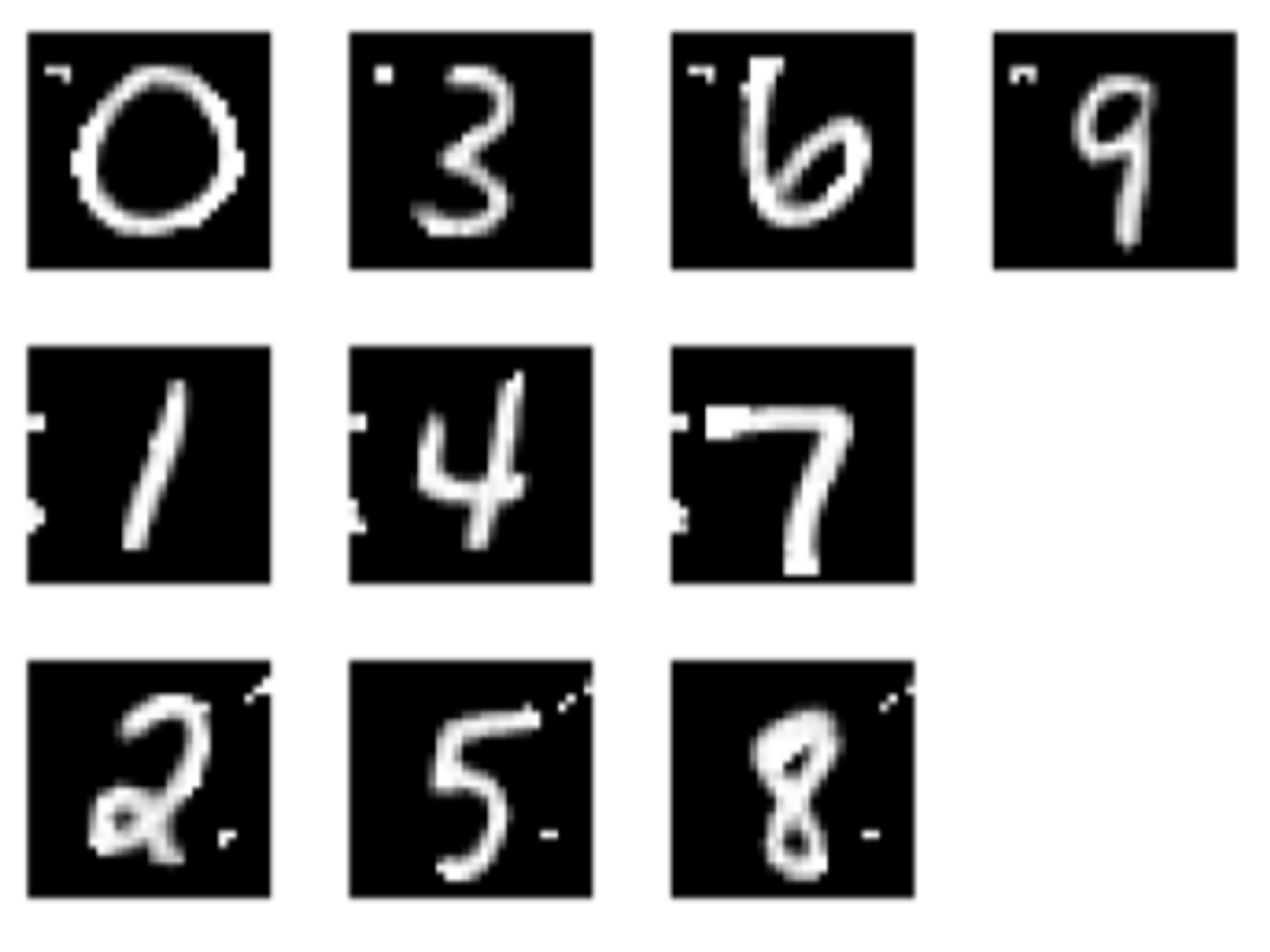}
  \caption{Reconstructed digits after introducing evidence of three groups. This is a case where the evidence introduced is the relation $y \mod 3$, with $y$ being the digit label. For visualization purposes, we move $Q$ estimators after the $X'$ reconstruction and use the Adam optimizer to clearly showcase the ``tagging'' of samples. For $y \mod 3 = 0$ there is a common pattern of marking the top left corner of the digit. The pattern of drawing two dots on the left side of the digit is deployed for samples where $y \mod 3 = 1$. Two shapes drawn in the right of the digit is deployed for $y \mod 3 = 2$ }
  \label{fig:digits}
\end{figure}

During the incremental manipulation of the initial latent space, latent representations are separated according to corresponding evidence. In order to minimize the joint loss objective, the learned representations are being manipulated into representing the evidence in their features. In Figure \ref{fig:digits} we change the position of predictors $Q$ in order to visually showcase the evidence being transferred in the latent representations. Changing the position of the additional layers during real corresponding evidence results to producing ``tagged'' samples. The ``tagging'' refers to data samples being reconstructed with added symbols that are consistent in the same way as the groupings of the evidence.

\begin{figure*}
  \centering
  \subfigure[Initial state of latent space for CIFAR-10]{\includegraphics[width=0.42\textwidth]{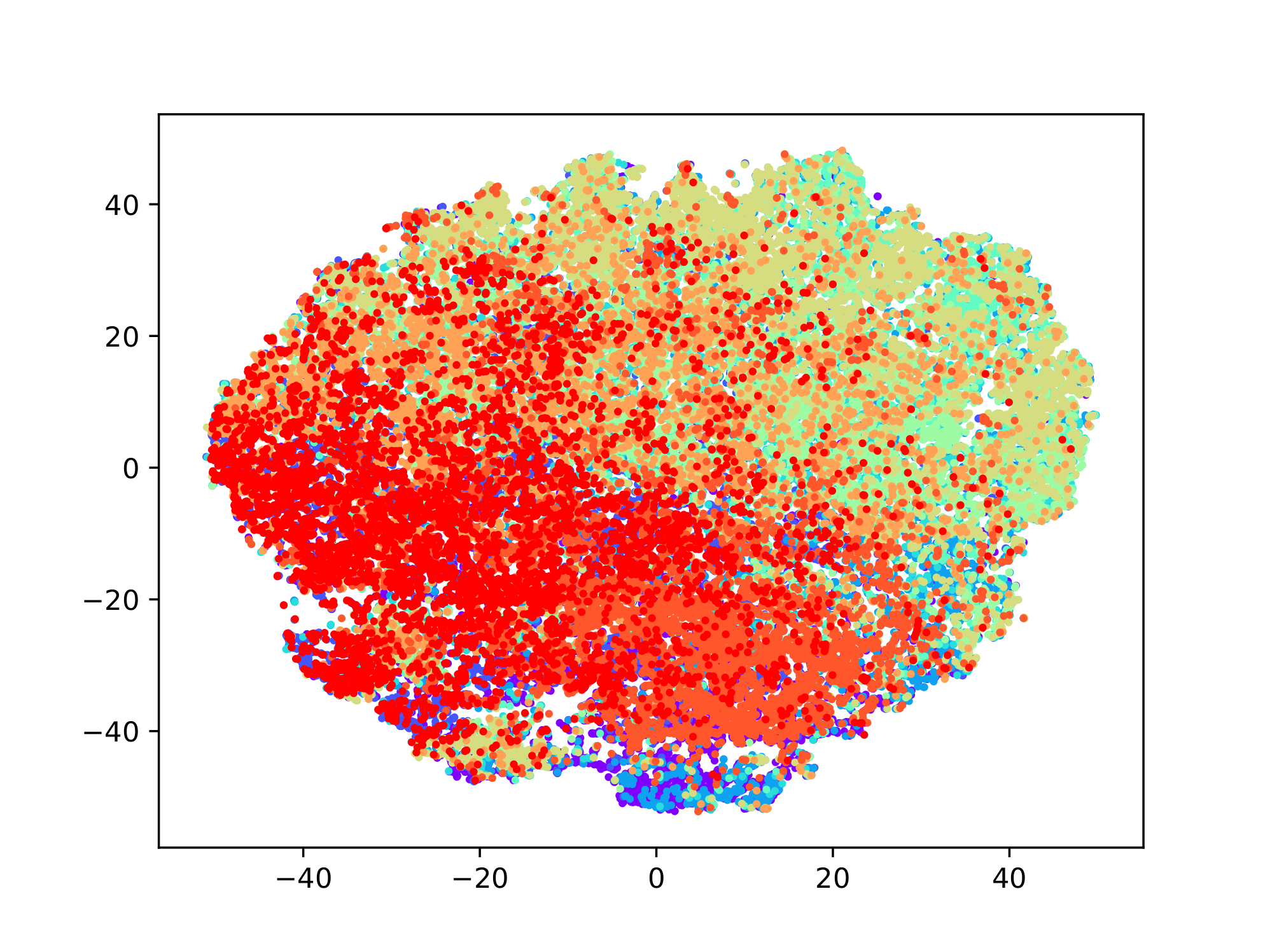}}
  \subfigure[Latent space of CIFAR-10, after incremental manipulation using real evidence (w:3)]{\includegraphics[width=0.42\textwidth]{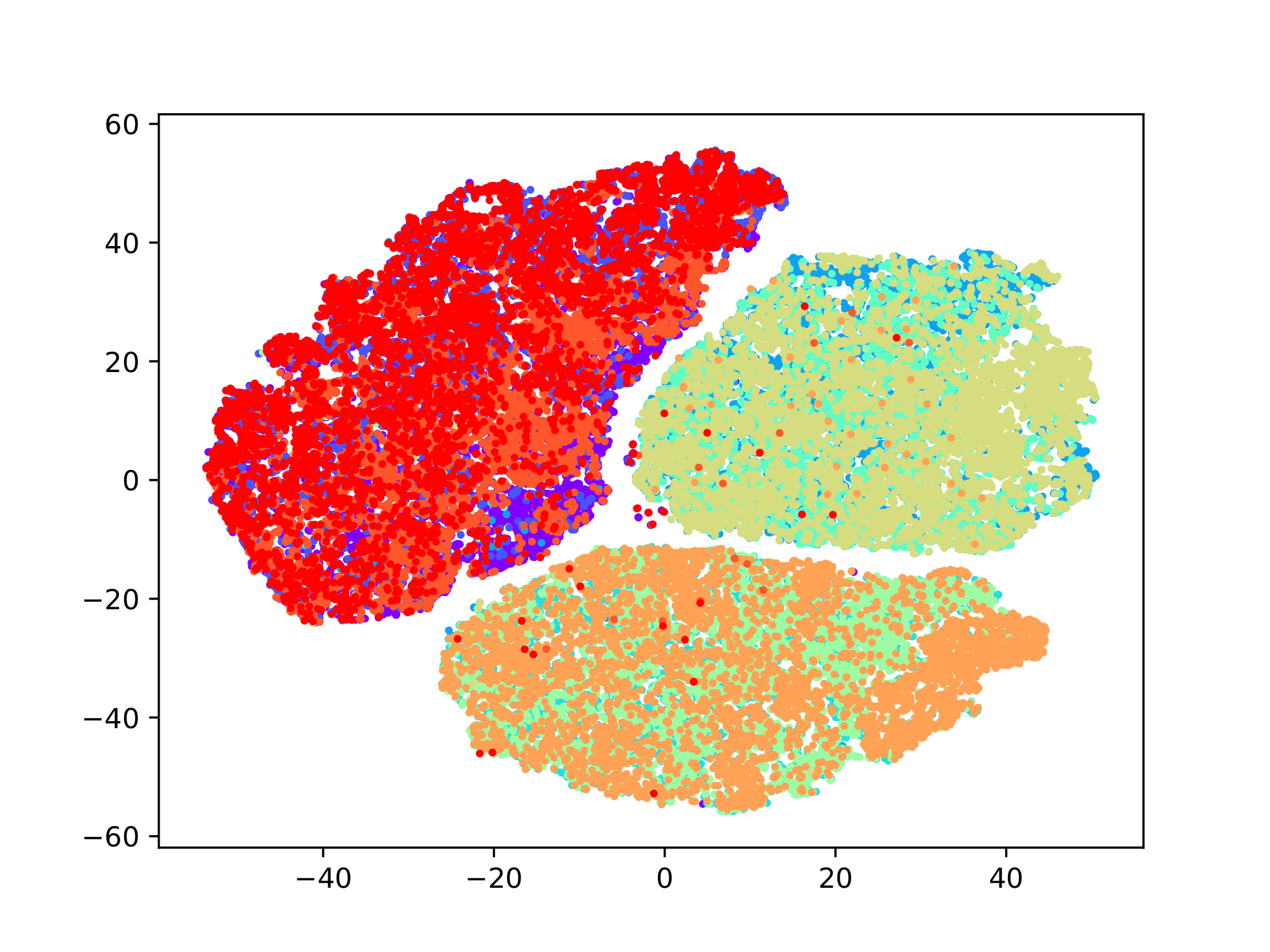}}
  \caption{We showcase the latent space of our primary autoencoder by transforming the latent representations into 2d samples, using t-SNE. Figure (a) depicts the initial state of the latent space for the CIFAR-10 dataset. Each latent sample is clustered around the mean. Figure (b) showcases the latent space after we introduced evidence of three groups (Real w:3). Incremental manipulation according to that evidence reforms the latent space into three distinct groups. They still preserve their initial structure, yet samples that indicate separation are completely separated.}
  \label{fig:cifar_lat}
\end{figure*}
\begin{figure*}
  \centering
  \subfigure[Initial latent representations of classes ``frog'' and ``automobile'']{\includegraphics[width=0.4\textwidth]{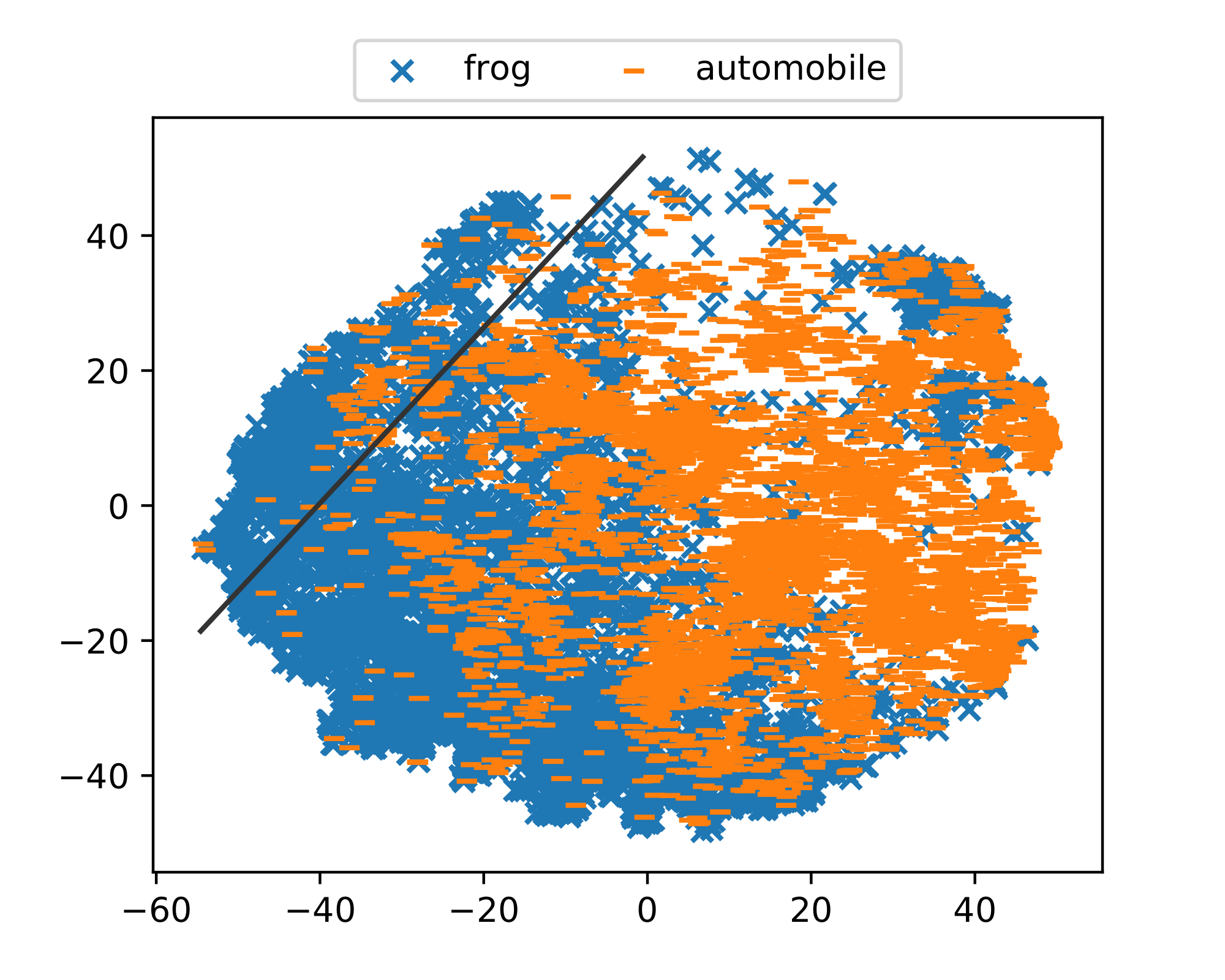}}
  \subfigure[Latent representations after introducing evidence that indicates separation]{\includegraphics[width=0.4\textwidth]{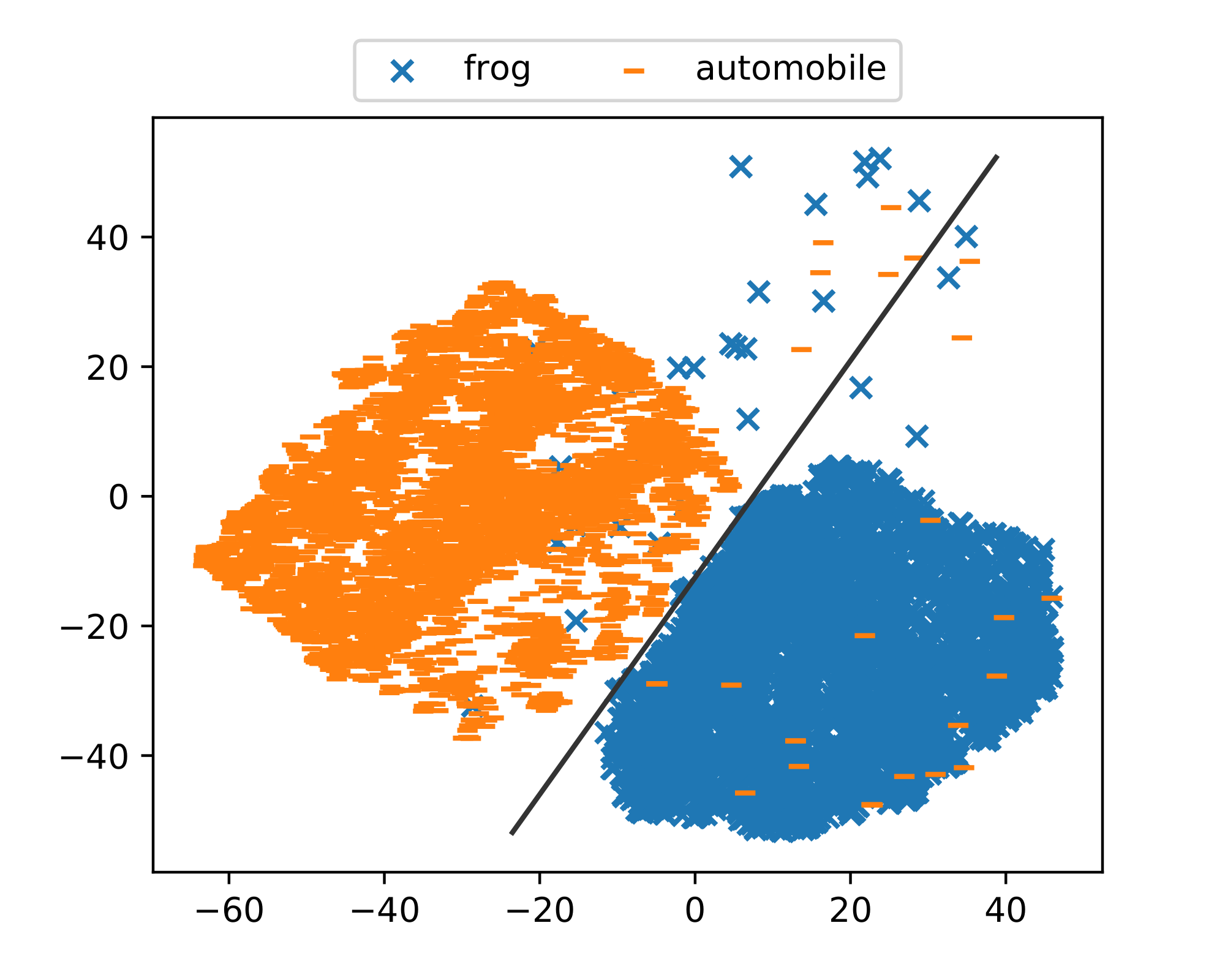}}
  \subfigure[Initial latent representations of classes ``frog'' and ``bird'']{\includegraphics[width=0.4\textwidth]{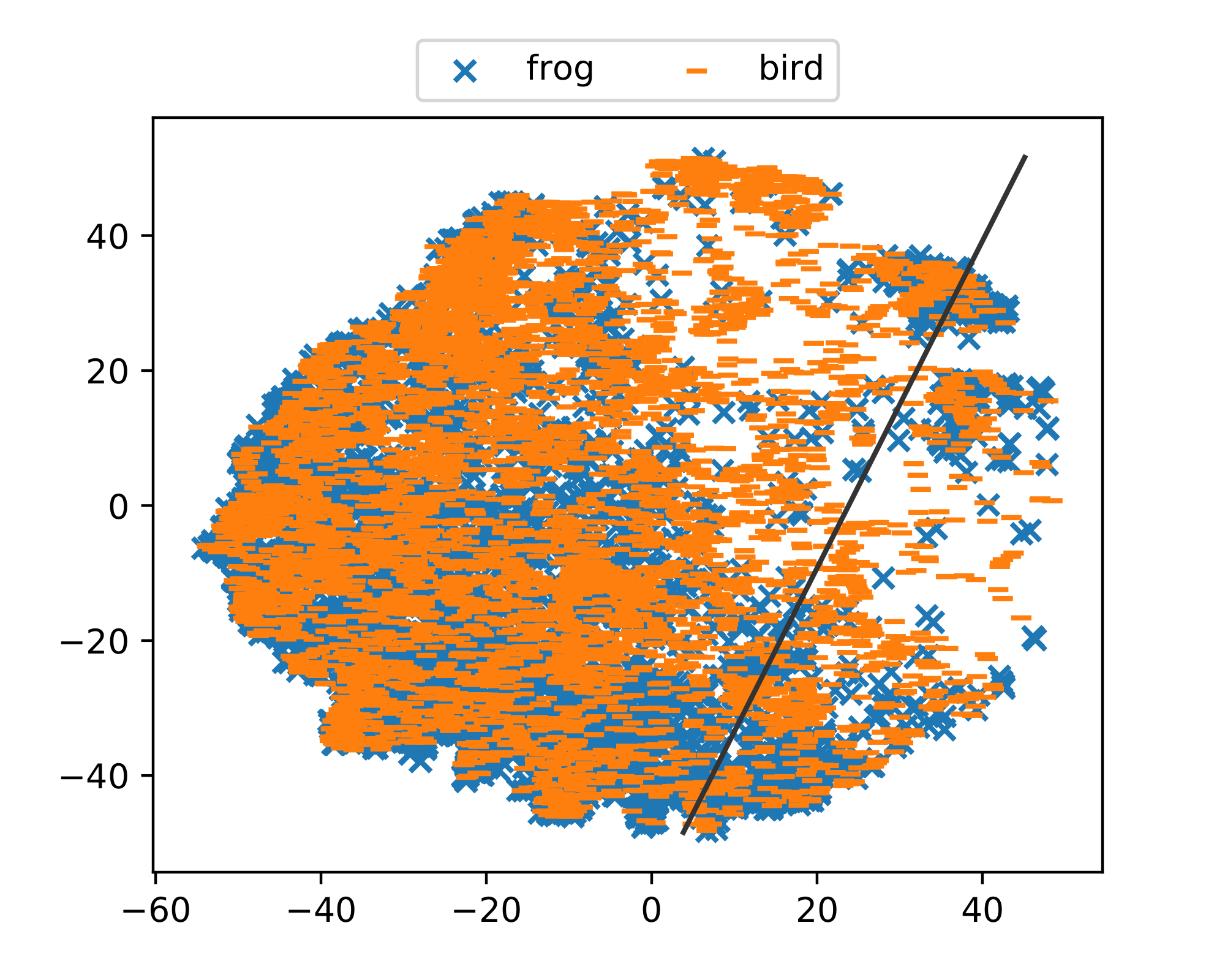}}
  \subfigure[Latent representations after introducing evidence that ``frog'' and ``bird'' should not be separated]{\includegraphics[width=0.4\textwidth]{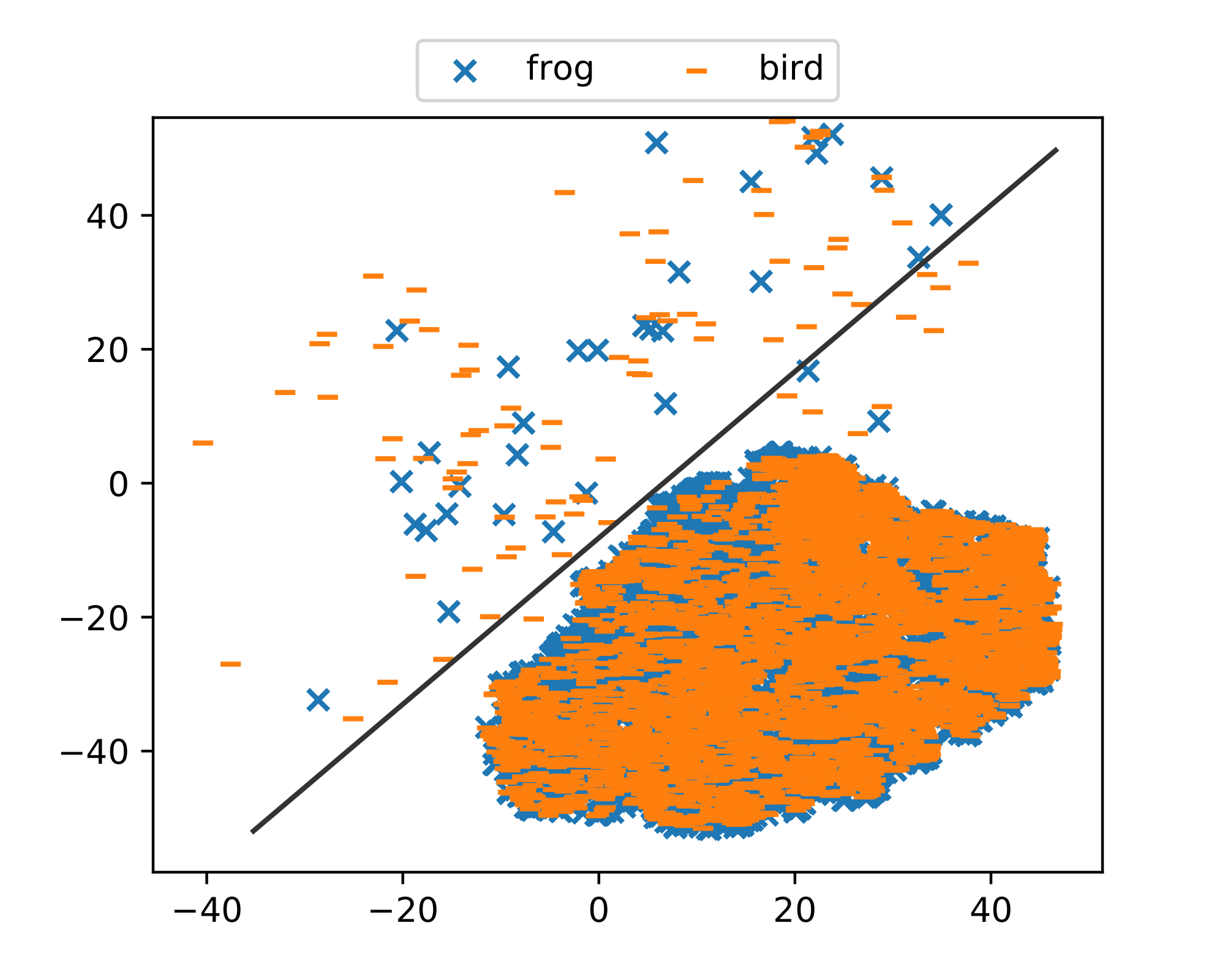}}
  \caption{In this figure we highlight the separations among the latent representations, by displaying the differences in comparing two labels. Straight lines indicate the decision boundary of an SVM classifier with linear kernel. For this experiment we used evidence that separates CIFAR-10 samples into 3 groups (Real w:3). Vehicles, pets and wild animals. According to our evidence, ``frog'' and ``automobile'' belong in different groups and therefore their latent representations should reflect their relation. On the other hand, ``frog'' and ``bird'' are both ``wild animals'' and their latent representations should indicate their shared group.}
  \label{fig:cifar_lat2}
\end{figure*}

To evaluate the ability of a linear algorithm to distinguish between samples of different classes after the incremental manipulation, we use an SVM classifier (with linear kernel) to evaluate the ability to perform binary classification before and the after the evidence transfer. Figure \ref{fig:cifar_lat2} showcases the ability of an SVM classifier to distinguish between two specific classes during the initial and incremental manipulation stage. Figure \ref{fig:cifar_lat} visualizes both states of latent space as a whole.

\paragraph{Reuters Optimization} For our experiments using the Reuters dataset, we incrementally trained the initial solution by alternating between minimizing \recon{} and \cross{} in each batch. This disjoint optimization of the two losses was deployed in order to ensure consistency in satisfying the robustness criterion. The initial latent space of our autoencoder consists of distinct small clusters with overlapping samples. These small clusters are a result of the most frequent word stems in some categories such as ``Employment/Labour'' (E41) associating with multiple other categories such as ``Domestic Politics'' (GPOL). Disjointly optimizing the two losses with \recon{} having higher learning rate than \cross{} more consistently was able to manipulate the latent space into having these initial properties.


\section{Related Work}
\label{sec:relwork}

\paragraph{Deep Generative Networks} CrossingNets \cite{Wan2017} is one of the frameworks that have considered combining external information in an unsupervised learning task. CrossingNets uses two different data sources for the task of hand pose estimation. It combines the two data sources by using a shared latent space. Although effective for the task of hand pose estimation, CrossingNets is a task specific configuration that does not propose a scalable way to handle multiple sources of evidence.

InfoGAN \cite{Chen2016} is also another case of using external information in the unsupervised task of generation. InfoGAN uses structured latent variables in order to manipulate the generation process of a GAN. InfoGAN utilizes the Information Maximization \cite{Barber2003} algorithm in order to disentangle latent representations. The structured latent variables $c$ are arbitrarily chosen by observing the dataset and are randomly sampled from known distributions. Latent variables $c$ are considered as low quality evidence in our case, since they do not provide any insight for the primary task.

\paragraph{Autoencoders} Adverarial Autoencoders (AAE) \cite{Makhzani2015} assimilate external information into the latent space, using random samples. AAE uses external information by making assumptions upon the prior distribution of the latent space. They depend on sampling from known distributions such as the Normal or Categorical distribution, to manipulate the latent space of an autoencoder. As with InfoGAN, types of evidence as such are considered as low quality and serve a specific purpose in these frameworks. In addition, AAE makes assumptions regarding the structure of the latent space and uses the external information accordingly.

Multi modal datasets have been used in autoencoders in order fill missing sensor data \cite{Jaques2017}. The latent space of the Multimodal Autoencoder (MMAE) has assimilated all modalities of a mood prediction dataset and is able to reconstruct images even in cases where some modalities are missing. MMAE is a specific configuration designed as a way to accommodate the cases of missing sensor data and proposes a way of assimilating modalities of the same dataset.

To solve imbalance in classification samples \cite{Ng2016} proposed a dual autoencoder solution. These autoencoders use different activations of the same autoencoder to capture different aspects of the same information. Each different configuration can be treated as an additional external information in the same way as ensemble methods combine different task outcomes. Another way of defining dual autoencoders configuration is for learning representations used for recommendation \cite{Zhuang2017}. These autoencoders focus on two different sources (namely item and user) in order to create a latent space that will be used for the recommendation task.

These methods make assumptions upon the availability of the external information. These autoencoder configurations require both datasets to be present during prediction of new data.

%

\paragraph{Semi-Supervised Learning} In Semi-supervised learning the various configurations make use of both labeled and unlabeled data. The presence of labels or lack thereof usually refers directly to the labelset associated with the training dataset as well as the primary task. Semi-supervised learning configurations (\cite{Springenberg2015}, \cite{Dumoulin}, \cite{Belghazi}), Principled Hybrids of Generative and Discriminative \cite{Lasserre2006} and our framework, share the notion of training Generative models to also perform Discriminative tasks. Nevertheless, in our framework the external categorical evidence is not guaranteed to be directly associated with the primary task.

\paragraph{Transfer Learning and Style transfer} Evidence transfer is by definition a transfer learning configuration where the weights of a pre-trained autoencoder are trained to integrate the external evidence in the task of reconstruction. As defined in Learning Without Forgetting \cite{Li}, evidence transfer belongs in the ``Joint optimization'' methods where the weights of the pre-trained model are fine-tuned along with the randomly initialized weights of the auxiliary tasks. According to \cite{Li}, joint optimization does not forget the original task, while still having the best performance in the auxiliary task.

In style transfer \cite{Gatys2015} freezing the pre-trained layers of the network is the standard process. During style transfer the primary neural network does not update its trainable parameters. It reconstructs original images (content) using the style features of auxiliary images (style). Evidence transfer, produces evidence informed weights and biases that are used to produce latent representations. The incremental representation learning that we perform, increases the effectiveness in the clustering task. Using these weights we can encode and make predictions about new data with no expectations regarding the availability of the external evidence. 


\paragraph{Feature aggregation} Feature aggregation methods (\cite{Savenkov2017}, \cite{Park2016} and \cite{Li2015}) rely on combining external knowledge in the form of additional features. They use auxiliary and primary dataset features in a co-learning configuration in order to produce latent representations as a result of aggregating meaningful features. These methods, like the dual or multimodal autoencoders, require all external data to be present during evaluation. Evidence transfer is based on the notion that expecting external knowledge to be present when predicting new data is not realistic and therefore this assumption is removed from our solution.

\paragraph{Multi-task learning} Multi task learning frameworks such as \cite{Kendall}, rely on learning external tasks to produce meaningful learning representations using knowledge from all external tasks. Multi-task learning introduces the assumption that the external evidence is directly related to the task of the primary dataset and the effectiveness of the configuration directly relies on that relation. The expectation of only introducing meaningful external evidence is unrealistic, since the extend to which an auxiliary task might increase the effectiveness of a primary task is not clear in all cases. Our solution avoids such expectations and adapts its performance according to the quality of the external evidence.

\section{Conclusions and Future Work}
\label{sec:conc}

In this paper we presented the evidence transfer method. Evidence transfer is a general method of combining external additional evidence to increase the performance of clustering tasks. It makes no assumptions about the relation of the evidence to the primary dataset or its availability. We introduced a set of guidelines to cope with the categorical evidence, as well as the evaluation criteria of a solution that exploits one or more pieces of external evidence to improve a primary task. Our evidence transfer approach manipulates the latent representation to be more linearly separable and therefore leads to increased performance. 
The effectiveness achieved scales with increasing number of pieces of evidence, while at the same time it is robust when introducing low quality or ineffective evidence. 

We evaluated our proposed solution using evidence of different quantitative and qualitative properties, however effective evidence is directly related to our primary clustering task. Future work is directed to evaluating this method when there is a non-linear relationship between the evidence and the primary dataset. Although this case may be partially covered by using the latent representations of the additional evidence autoencoder, it remains to be validated experimentally. Additionally, evaluating the ability of evidence transfer to improve other unsupervised tasks such as generation is also considered for future work. 

Satisfying the robustness criterion during the optimization was a challenging task. The choice of optimizing algorithm proved to be crucial for the satisfaction of both the effectiveness and robustness, adaptive optimizers proved to disrupt the initial latent space during joint training with low quality of evidence. These experiments indicate further investigation of optimization techniques deployed on incremental training of multi task learning models with conflicting or unrelated objectives. 

\section*{Acknowledgments}
This work has been supported by the Industrial Scholarships program of Stavros Niarchos Foundation.

\bibliographystyle{IEEEtran}
\bibliography{IEEEabrv,evidence}

\end{document}